\documentclass[10pt,twocolumn,letterpaper]{article}

\usepackage[pagenumbers]{cvpr}

\def \customparskip {0.7em}
\renewcommand{\paragraph}[1]{\vspace{\customparskip}\noindent\textbf{#1}}

\newcommand{\suppmat}{\texttt{supplementary appendix}\xspace}

\newcommand{\tabvenue}[1]{%
  \hspace{2pt}{\color{gray}\fontsize{5.7}{6.8}\selectfont (#1)}%
}

\definecolor{cvprblue}{rgb}{0.21,0.49,0.74}
\usepackage[pagebackref,breaklinks,colorlinks,allcolors=cvprblue]{hyperref}

\title{BreathSplat: Grounding One Smooth Motion in the \\ Breathing Phase for 4D Cone-Beam CT}

\author{Sujong Shin$^{1}$ \qquad Sang-Il Choi$^{1}$ \qquad Kwang Moo Yi$^{2}$\\
$^{1}$Dankook University \qquad $^{2}$University of British Columbia\\
{\tt\small shinsjn@dankook.ac.kr \quad choisi@dankook.ac.kr \quad kmyi@cs.ubc.ca}
}

\begin{document}
\maketitle
\begin{abstract}
Four-dimensional cone-beam computed tomography (4D-CBCT) images the anatomy of a breathing patient, but sees each breathing phase from only a few angles.
Breathing, however, is a smooth motion that repeats---it has a phase.
Existing methods exploit this either by reconstructing the anatomy as a function of the breathing phase, often through bins, or by estimating smooth dynamic motions in time.
Yet none of these grounds one smooth motion in the phase.
We thus propose to share projections across breathing phases with splines.
Specifically, we represent the 4D volume with dynamic Gaussians whose motion uses B-splines in space and time.
We ground the motion state in the respiratory phase, which we read from well-studied conventional surrogates.
Projections at the same phase in different breathing cycles then constrain the same anatomy and motion.
On simulated lung scans, our method, BreathSplat, outperforms recent dynamic methods in reconstruction and motion accuracy, including those given the ground-truth phase.
We further show qualitatively, on two real patient scans, that our method provides clearer reconstructions than the state of the art.
\end{abstract}

\section{Introduction}
\label{sec:intro}

\begin{figure}[t]
\centering
\setlength{\tabcolsep}{1pt}%
\begin{tabular}{cc}
\includegraphics[width=0.48\linewidth]{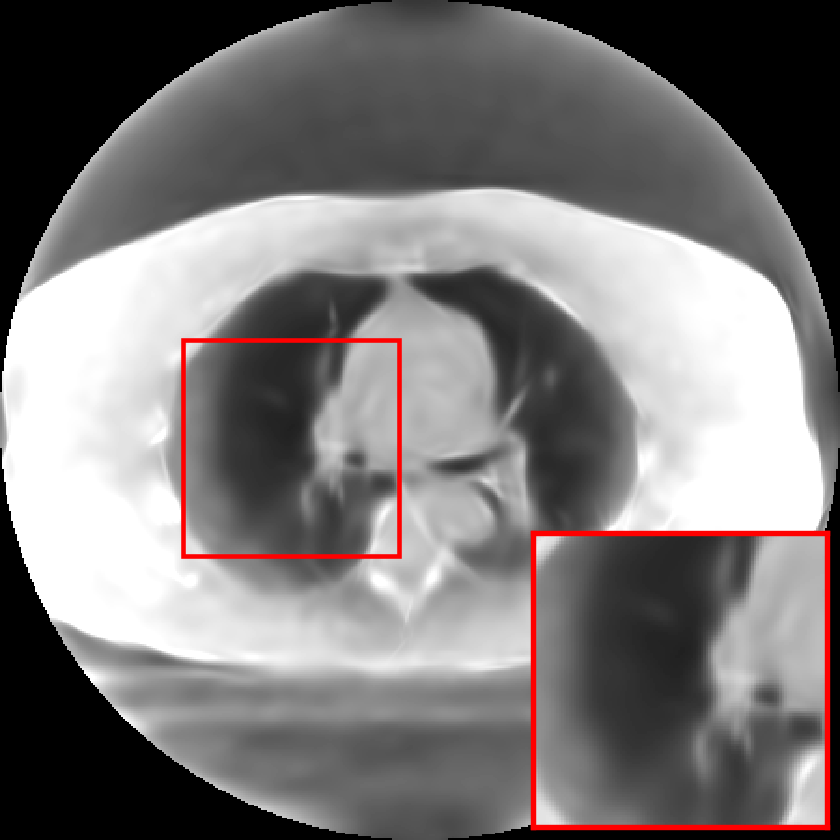} & \includegraphics[width=0.48\linewidth]{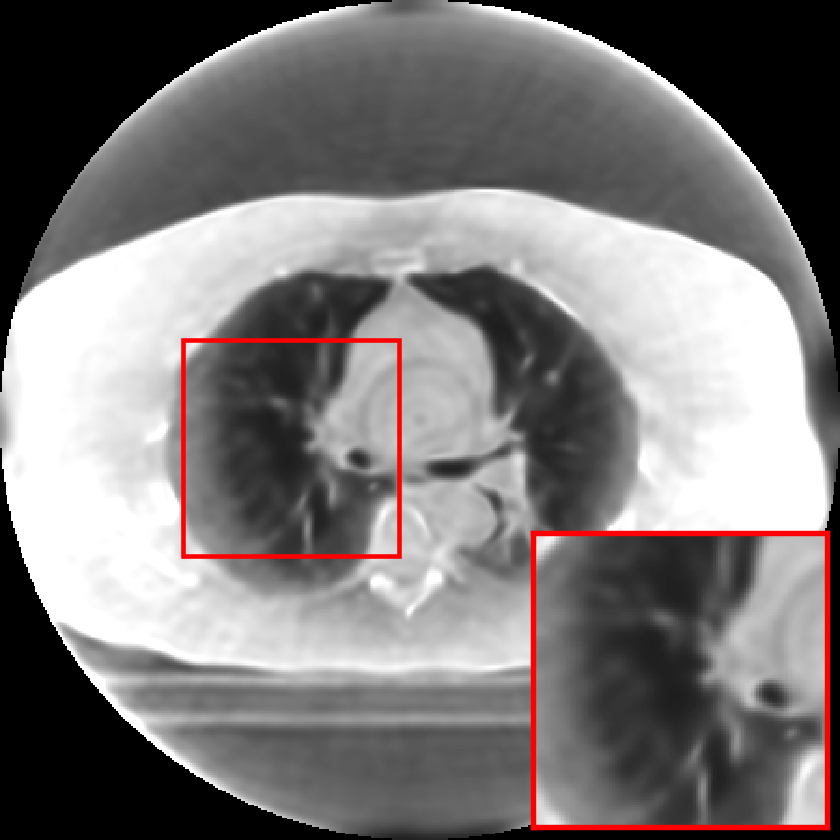} \\
\footnotesize X$^2$-Gaussian++~\cite{x2gaussianpp2026code} & \footnotesize Fu \etal~\cite{fu2025dynamiccbct} \\[3pt]
\includegraphics[width=0.48\linewidth]{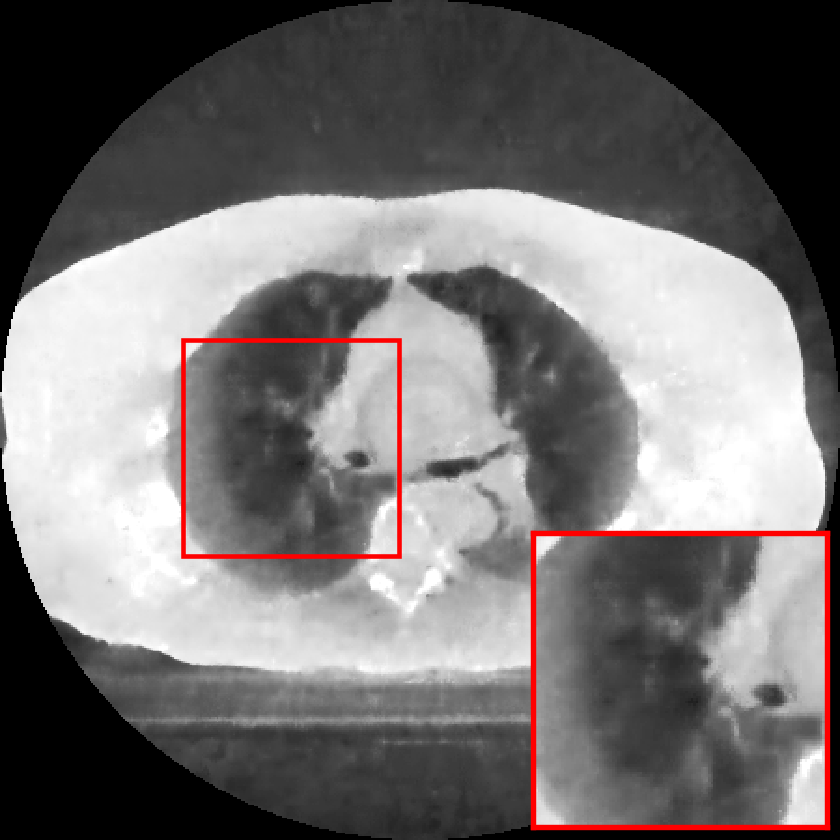} & \includegraphics[width=0.48\linewidth]{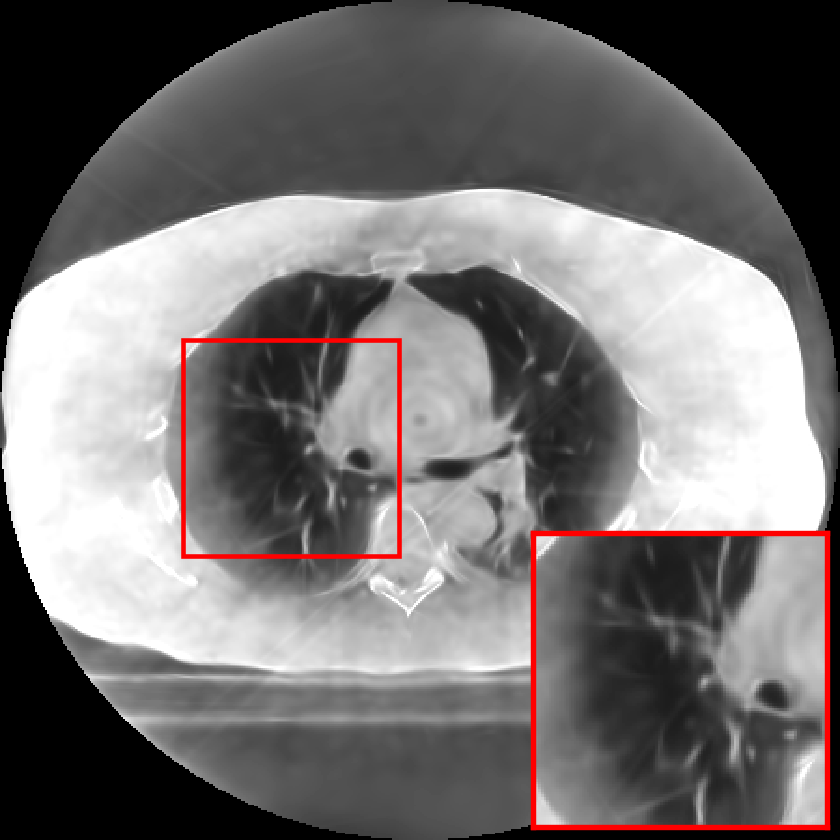} \\
\footnotesize STNF4D~\cite{zhou2025stnf4d} & \footnotesize \textbf{BreathSplat (ours)} \\
\end{tabular}
\caption{\textbf{Teaser.}
Our method reconstructs a real patient scan more sharply than the state of the art.
Pulmonary vessels and bone edges that the dynamic baselines blur stay sharp in our reconstruction.
Each panel shows the same axial slice of SPARE~\cite{shieh2019spare} (Varian patient 3, phase bin 01) with the same display window, and the inset enlarges the marked region.
}
\label{fig:teaser}
\end{figure}

Lung cancer kills more people than any other cancer~\cite{bray2024global}.
Radiotherapy can destroy a lung tumor, but the tumor moves with every breath, so the radiation must hit a moving target while sparing the healthy lung around it~\cite{sonke2009frameless}.
Four-dimensional cone-beam computed tomography (4D-CBCT) lets clinicians see this moving target right before each treatment, so that they can aim the radiation at it with tight margins~\cite{sonke2005respiratory,sonke2009frameless}.
It reconstructs the breathing anatomy of a patient from the X-ray projections of one gantry rotation.
Because the gantry keeps rotating while the patient breathes, the scanner sees each breathing phase from only a few angles~\cite{sonke2005respiratory}.
Each phase is thus severely undersampled, and reconstructing it from its own projections alone leaves streak artifacts~\cite{obrien2014optimizing}.
This, however, can be mitigated since breathing is a smooth motion that repeats, \ie, it has a phase.

Existing 4D-CBCT methods thus take advantage of this~\cite{sonke2005respiratory,li2007enhanced,yu2025x2gaussian,huang2025digs}.
Some ground the anatomy in the breathing cycle, so that the anatomy is the same whenever breathing returns to the same state.
Among these methods, most ground it in discrete phase bins.
They reconstruct each bin from its own projections~\cite{sonke2005respiratory}, share projections between adjacent bins~\cite{obrien2014optimizing}, deform one anatomy with a motion for each bin~\cite{li2007enhanced,wang2013smeir,brehm2012selfadapting,fu2025dynamiccbct}, or represent the 4D volume with a neural field that takes a 3D point and the bin index as input and directly outputs the attenuation at that point~\cite{zhou2025stnf4d}.
D-NAF~\cite{zhang2026dnaf} uses such a neural field, but with a smooth, periodic encoding of the phase instead of the bin index.
Others ground it in acquisition time, by asking the anatomy to repeat after one learned breathing period~\cite{yu2025x2gaussian,x2gaussianpp2026code}.
Methods that reconstruct each bin~\cite{sonke2005respiratory,obrien2014optimizing} or estimate a motion for each bin~\cite{li2007enhanced,wang2013smeir,brehm2012selfadapting,fu2025dynamiccbct}, however, fit the image or motion of each bin mainly to the few projections at or near its phase.
Neural fields that output the attenuation directly~\cite{zhou2025stnf4d,zhang2026dnaf} deform no shared anatomy, so a projection informs the other phases only through the shared network weights.
A learned period~\cite{yu2025x2gaussian,x2gaussianpp2026code}, in turn, only encourages the anatomy to repeat, and assumes every breathing cycle lasts equally long.

Other methods~\cite{shao2024pmfstinr,xie2025pmfstgr,huang2025digs} focus on modeling the breathing motion itself, with neural fields or dynamic extensions~\cite{yang2024deformable,wu20244dgs} of 3D Gaussian splatting (3DGS)~\cite{kerbl2023gaussian} that reconstruct the anatomy at every instant of the scan.
Neural fields~\cite{shao2024pmfstinr} and the low-rank B-spline motion of one set of Gaussians in DIGS~\cite{huang2025digs} model the motion as a smooth function of \emph{acquisition time}.
Even aDiner~\cite{zhou2026adiner}, which reads the phase from a surrogate, conditions its motion on the index of the breathing cycle, effectively going back to acquisition time.
A motion on acquisition time~\cite{shao2024pmfstinr,huang2025digs,zhou2026adiner} or on each projection~\cite{xie2025pmfstgr}, however, need not repeat with the breathing.
To the best of our knowledge, no 4D-CBCT reconstruction method learns one smooth motion of one anatomy as a function of the respiratory phase directly from the scan's projections.
As we show in \cref{sec:ablations}, doing so is important for the best reconstruction.

Our core idea is to share projections across breathing \emph{phases} with \emph{B-splines}.
We deform one set of Gaussians, a single anatomy that all phases share, with one motion that uses B-splines in space and time.
To capture the physical properties of breathing, we define the temporal B-splines on the respiratory phase, which is periodic, and not directly on time.
Conventional surrogate methods~\cite{sonke2005respiratory,hugo2012fourdimaging,yan2013extracting} already read this phase from the projections.
In more detail, we propose to combine a conventional breathing signal read from the projections~\cite{yan2013extracting} with an established model of breathing traces whose rate varies over time~\cite{ruan2009realtime}.
With this model, projections at the same phase in different breathing cycles constrain the same anatomy and motion, each from a different gantry angle, making the problem more constrained.
As all phases also share this anatomy and the spatial B-splines of this motion, every projection constrains both, whatever its phase.
The B-splines on the phase, periodic as in four-dimensional CT (4DCT) registration~\cite{metz2011nonrigid,vandemeulebroucke2011spatiotemporal}, keep the motion smooth.
In short, every projection informs the smoothly moving anatomy at every phase.

To demonstrate the efficacy of our method, we evaluate it quantitatively and qualitatively on simulated scans, and qualitatively on two real, publicly available patient scans.
On scans with irregular breathing that we simulate from public lung datasets, our method, BreathSplat, outperforms the state of the art.
Our method reaches a body-volume peak signal-to-noise ratio (PSNR) of 42.18\,dB and a motion error of 0.57\,mm, which is respectively 2.78\,dB and 0.40\,mm better than the second-best method in each metric (\cref{tab:main-results}).
It also tracks expert-annotated landmarks best (\cref{tab:tre}).
We further show that a single smooth motion describing the breathing provides the best results.
On two real patient scans, BreathSplat shows fine structures more clearly than the dynamic baselines~(\cref{fig:teaser,fig:clinical}).

To summarize, our contributions are:
\begin{itemize}
    \item we propose to share projections across breathing phases with splines, by grounding one B-spline motion of one set of Gaussians in the respiratory phase that a conventional surrogate reads from the projections;
    \item we provide a thorough ablation study of where our gains come from, which shows that grounding the motion in the phase is important; and
    \item we significantly outperform the state of the art, demonstrated quantitatively and qualitatively by synthetic experiments and qualitatively by real patient scans.
\end{itemize}

\section{Related work}
\label{sec:related}

\vspace{-\customparskip}
\paragraph{Conventional 4D-CBCT.}
Conventional 4D-CBCT shares projections across breathing phases in several ways.
Respiratory-correlated CBCT~\cite{sonke2005respiratory} sorts the projections into phase bins and reconstructs each bin, with Feldkamp--Davis--Kress (FDK) reconstruction~\cite{feldkamp1984practical} or iterative methods~\cite{sidky2008image}.
Bin sharing~\cite{obrien2014optimizing} lets adjacent bins share projections, and prior image constrained compressed sensing (PICCS)~\cite{chen2008piccs} constrains each bin with a prior image from all projections.
Motion-compensated methods reconstruct one image of the anatomy, and compensate a motion that they estimate for each bin~\cite{li2007enhanced,wang2013smeir,brehm2012selfadapting} or retrieve from a prior 4DCT scan~\cite{rit2009motioncompensated}.
The model of Huang \etal~\cite{huang2024surrogate}, the closest to ours, instead makes the motion a function of surrogate values.
Methods that reconstruct each bin~\cite{sonke2005respiratory,obrien2014optimizing} or estimate a motion for each bin~\cite{li2007enhanced,wang2013smeir,brehm2012selfadapting}, however, fit the image or motion of each bin mainly to the few projections at or near its phase.
Surrogate values~\cite{huang2024surrogate} can follow changes in breathing amplitude, but DIGS~\cite{huang2025digs} reports this model to be less accurate than learned dynamic reconstruction.
Unlike these methods, we share projections through one motion of one anatomy, defined on a continuous phase and fit to the projections of the scan itself.

\paragraph{Projection-based respiratory surrogates.}
Projection-based surrogates~\cite{sonke2005respiratory,hugo2012fourdimaging,yan2013extracting} read the breathing signal from the projections themselves, without external hardware, by separating respiration-induced changes from those that the gantry rotation causes.
As we show, when combined with dynamic motion models, they provide a powerful tool for describing breathing motion.
We adapt the surrogate of Yan \etal~\cite{yan2013extracting}, and fit a repeating waveform to its signal to obtain a continuous phase (\cref{sec:respiratory-timing}).

\paragraph{Learned dynamic reconstruction.}
Learned methods improve 4D-CBCT in two ways.
Supervised networks~\cite{jiang2019augmentation,zhi2021cycnnet} learn from many scans to restore undersampled reconstructions, such as those of phase-binned 4D-CBCT~\cite{zhi2021cycnnet}.
Scan-specific methods~\cite{shao2024pmfstinr,yu2025x2gaussian,huang2025digs} instead fit a dynamic representation to the projections of one scan alone.
They adapt static neural attenuation fields~\cite{zha2022naf} and radiative Gaussians~\cite{zha2024r2gaussian} to the breathing anatomy, as well as deformable 3DGS~\cite{yang2024deformable,wu20244dgs}, which reconstructs dynamic natural scenes by moving one set of Gaussians with a learned deformation field.
Supervised networks~\cite{jiang2019augmentation,zhi2021cycnnet}, however, are only as accurate as their training data cover the anatomy, motion, and imaging parameters of a new scan~\cite{shao2024pmfstinr}.
Like scan-specific methods, we thus fit each scan on its own.

Scan-specific methods vary according to what they define the 4D volume on.
Many define it on acquisition time, with neural fields~\cite{zhang2023stinr,shao2024pmfstinr,zhong2026tgfield}, feature planes~\cite{fridovichkeil2023kplanes,cao2023hexplane} with a deformation decoder~\cite{yu2025x2gaussian,x2gaussianpp2026code}, or low-rank B-splines~\cite{huang2025digs}, infer it from each projection~\cite{xie2025pmfstgr}, or pair a surrogate phase with the index of the breathing cycle~\cite{zhou2026adiner}.
These, however, are not made to repeat with the breathing.
A periodic consistency loss~\cite{yu2025x2gaussian,x2gaussianpp2026code}, in turn, only encourages them to.
Others take the phase from outside the reconstruction, as discrete phase indices~\cite{fu2025dynamiccbct,zhou2025stnf4d} or a periodic encoding of the phase~\cite{zhang2026dnaf}.
Discrete phase indices~\cite{fu2025dynamiccbct,zhou2025stnf4d} tie each projection to one phase, even when a loss~\cite{fu2025code} encourages smoothness across them.
STNF4D~\cite{zhou2025stnf4d} and D-NAF~\cite{zhang2026dnaf} predict the attenuation directly and deform no shared anatomy.
Fu \etal~\cite{fu2025dynamiccbct} deform one anatomy, but also change the density of its Gaussians at each phase.
We build our motion on the low-rank B-spline motion of DIGS~\cite{huang2025digs}, but define its temporal B-splines on the respiratory phase and make them periodic, so that projections at the same phase in different breathing cycles constrain the same anatomy and motion.

\paragraph{Periodic B-spline motion in 4DCT registration.}
Spatiotemporal registration for 4DCT~\cite{metz2011nonrigid,vandemeulebroucke2011spatiotemporal} models breathing with B-spline free-form deformations~\cite{rueckert1999nonrigid} that are also B-splines in time, periodic over the breathing cycle.
These methods register reconstructed 4DCT images, whereas we fit such a motion directly to sparse CBCT projections.

\section{Method}
\label{sec:method}

\subsection{Problem formulation}
\label{sec:problem-formulation}

\begin{figure*}[t]
\centering
\includegraphics[width=\textwidth,trim=31.1073bp 40.6318bp 31.8667bp 61.3661bp,clip]{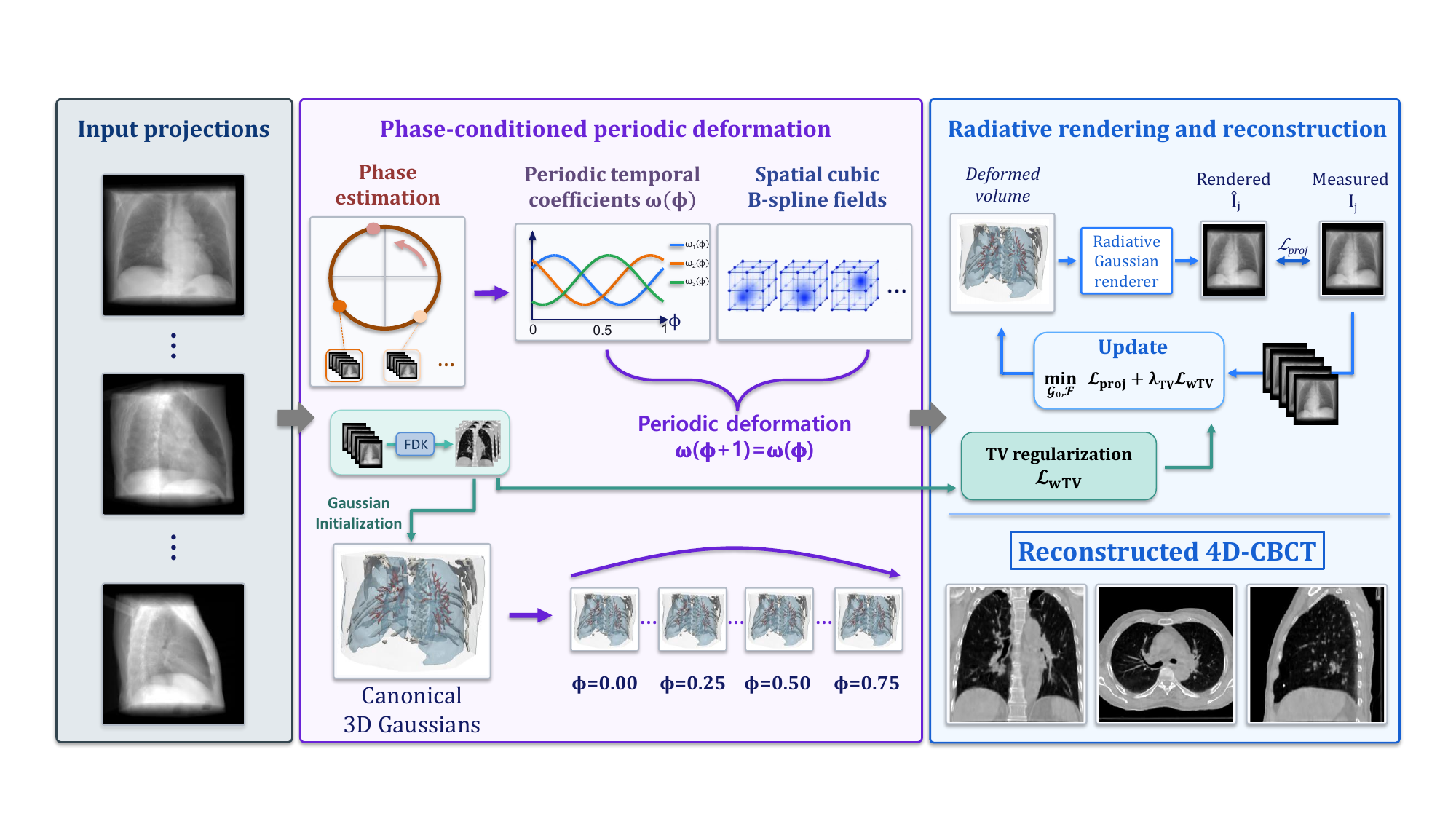}
\vspace{-2em}
\caption{\textbf{Overview.}
Before optimization we use a conventional surrogate to retrieve the respiratory phase of every projection.
We deform one set of canonical Gaussians with the low-rank B-spline motion of DIGS~\cite{huang2025digs}, whose temporal B-splines we define on this phase, making the motion periodic and aligned to breathing cycles by construction.
We then fit both the motion and the Gaussians with the projections, with a TV prior considering the edges.
}

\label{fig:method-overview}
\end{figure*}

Our input is a set of training projections $\{(I_j,A_j,t_j)\}_{j=1}^{N_{\mathrm{tr}}}$, where $I_j$ is a line-integral image, $A_j$ is its cone-beam geometry, and $t_j\in[0,1)$ is its acquisition time normalized by the scan duration.
We represent the anatomy with canonical radiative Gaussians $\mathcal{G}_0$, each with a center, scale, rotation, and density.
We render their X-ray projections with the differentiable renderer $\mathcal{R}$ of R$^2$-Gaussian~\cite{zha2024r2gaussian}.
To model breathing, we deform the Gaussians with a function $\mathcal{F}_{\phi}$~(\cref{sec:periodic-ffd}) of the respiratory phase $\phi\in[0,1)$, which a surrogate gives every projection $j$ as $\phi(t_j)$ (\cref{sec:respiratory-timing}).
We then solve, as \cref{fig:method-overview} summarizes,
\begin{equation}
\begin{split}
\min_{\mathcal{G}_0,\,\mathcal{F}}\;\frac{1}{N_{\mathrm{tr}}}\sum_{j}\Big[&\mathcal{L}_{\mathrm{proj}}\Big(\mathcal{R}\big(A_j,\mathcal{F}_{\phi(t_j)}(\mathcal{G}_0)\big),I_j\Big)\\
&+\lambda_{\mathrm{TV}}\mathcal{L}_{\mathrm{wTV}}(V_{\phi(t_j)})\Big],
\end{split}
\label{eq:objective}
\end{equation}
where $\mathcal{L}_{\mathrm{proj}}$ is a loss between rendered and measured projections, and $\mathcal{L}_{\mathrm{wTV}}$, with weight $\lambda_{\mathrm{TV}}$, is an edge-weighted total variation (TV) prior on the volume $V_{\phi(t_j)}$ that we voxelize~\cite{zha2024r2gaussian} from the Gaussians at the phase of projection $j$.

\paragraph{Reconstruction loss -- $\mathcal{L}_{\mathrm{proj}}$.}
To compare a rendered projection $\hat{I}$ with the measured projection $I$, we follow X$^2$-Gaussian++~\cite{yu2025x2gaussian,x2gaussianpp2026code} and combine an $\ell_1$ loss, a structural similarity (SSIM) loss~\cite{wang2004ssim}, and a wavelet loss~\cite{x2gaussianpp2026code}:
\begin{equation}
\begin{split}
\mathcal{L}_{\mathrm{proj}}(\hat{I},I)={}&\|\hat{I}-I\|_1+\lambda_{\mathrm{s}}\big(1-\mathrm{SSIM}(\hat{I},I)\big)\\
&+\lambda_{\mathrm{w}}\|W(\hat{I})-W(I)\|_1,
\end{split}
\label{eq:proj-loss}
\end{equation}
where $\|\cdot\|_1$ is the per-pixel $\ell_1$ norm, $W$ is a Haar wavelet high-pass filter, and $\lambda_{\mathrm{s}}$ and $\lambda_{\mathrm{w}}$ are hyperparameters that control the influence of the last two terms.

\paragraph{Total variation loss -- $\mathcal{L}_{\mathrm{wTV}}$.}
To suppress sparse-view artifacts with TV~\cite{rudin1992tv,sidky2008image} without blurring anatomical boundaries, we weaken TV on edges~\cite{tian2011edge} with voxel weights,
\begin{equation}
\mathcal{L}_{\mathrm{wTV}}(V)=\frac{\sum_{(p,q)}\min(w_p,w_q)\,\big|V(p)-V(q)\big|}{\sum_{(p,q)}\min(w_p,w_q)},
\label{eq:wtv}
\end{equation}
where $(p,q)$ is a pair of neighboring voxels, and $w_p$ is the weight of voxel $p$, which is lower on edges (\cref{sec:projection-priors}).

\subsection{Respiratory phase from projections}
\label{sec:respiratory-timing}

While X$^2$-Gaussian~\cite{yu2025x2gaussian,x2gaussianpp2026code} learns the breathing period jointly with the reconstruction, conventional respiratory surrogates~\cite{sonke2005respiratory,hugo2012fourdimaging,yan2013extracting} already read the breathing from the projections themselves---we thus use them.
We propose to combine two established tools to obtain a phase that advances continuously with the breathing.

Like Yan \etal~\cite{yan2013extracting}, we extract a breathing signal from the cranio-caudal intensity profiles of the training projections with principal component analysis (PCA).
Gantry rotation also changes these profiles~\cite{yan2013extracting}, because the X-rays cross the body along a different path at each gantry angle.
This change is a smooth, periodic function of the gantry angle, so we regress out harmonics of the angle, together with a slow temporal trend, before PCA.
We then fit to this signal a repeating waveform of a cumulative phase $\theta(t)$, plus a slowly drifting baseline~\cite{ruan2009realtime}.

Unlike the existing general warping formula~\cite{ruan2009realtime}, we let this phase advance at the mean breathing frequency $c$, and add a correction $\delta(t)$, a cubic spline with a curvature penalty~\cite{reinsch1967smoothing}, so that the phase can advance faster or slower as the breathing rate changes.
We thus write the phase of a projection acquired at time $t$ as
\begin{equation}
\phi(t)=\operatorname{frac}\big(\theta(t)\big),\qquad \theta(t)=ct+\delta(t),
\label{eq:phase-coordinate}
\end{equation}
where $\operatorname{frac}$ takes the fractional part.
As we show in \cref{tab:main-results}, our phase estimate is precise---replacing this phase with the ground-truth phase changes our accuracy little.
We thus keep the phase fixed during reconstruction.

\subsection{Breathing motion as splines on the phase}
\label{sec:periodic-ffd}
\label{sec:reconstruction-framework}

To share the projections across phases, we deform the canonical Gaussians with one motion that depends on time only through the respiratory phase $\phi$, and that is smooth and periodic in $\phi$.
We build this motion on the low-rank B-spline motion of DIGS~\cite{huang2025digs}, which factorizes it into $R$ spatial fields $\mathbf{u}_r$ and temporal coefficients $\omega_r$.
Each spatial field interpolates control-point offsets on a coarse lattice with a cubic B-spline free-form deformation~\cite{rueckert1999nonrigid}, which keeps the motion smooth in space.
DIGS moves the center of each Gaussian with this motion, and transforms its covariance with the Jacobian of the motion at the center, so that the shape of the Gaussian follows how the motion locally stretches and rotates the anatomy.
This keeps the shapes consistent with the motion, but does not let them change across phases beyond what the motion of the centers dictates.
We instead let the motion change the center, scale, and rotation of each Gaussian separately, as the deformation network of X$^2$-Gaussian~\cite{yu2025x2gaussian} does.

\paragraph{Low-rank deformation in space and phase.}
In more detail, let us define the geometry of Gaussian $n$ as $\mathbf{a}_n=(\boldsymbol{\mu}_n,\mathbf{s}_n,\mathbf{q}_n)\in\mathbb{R}^{10}$, where $\boldsymbol{\mu}_n\in\mathbb{R}^3$ is its center, $\mathbf{s}_n\in\mathbb{R}^3$ is its log-scale, and $\mathbf{q}_n\in\mathbb{R}^4$ is its rotation quaternion.
We further write its density as $\rho_n$, so that $\mathcal{G}_0=\{(\mathbf{a}_n,\rho_n)\}_{n=1}^{N_{\mathrm{G}}}$, where $N_{\mathrm{G}}$ is the number of Gaussians.
We then write the deformation $\mathcal{F}_{\phi}$ in \cref{eq:objective} as
\begin{equation}
\mathcal{F}_{\phi}(\mathcal{G}_0)=\big\{\big(\mathbf{a}_n+\Delta\mathbf{a}_n(\phi),\,\rho_n\big)\big\}_{n=1}^{N_{\mathrm{G}}},
\label{eq:deformation}
\end{equation}
where $\Delta\mathbf{a}_n(\phi)\in\mathbb{R}^{10}$ is the offset of Gaussian $n$ at phase $\phi$.
Thanks to the log scale and the quaternion parameterizations~\cite{kerbl2023gaussian}, we can add any offset without constraints.
As we want a motion that is smooth \emph{in space and in phase}, we write the offsets in a low-rank form, as a sum of $R$ B-spline spatial fields, each weighted by a temporal coefficient that we also define with a B-spline,
\begin{equation}
\Delta\mathbf{a}_n(\phi)=\sum_{r=1}^{R}\omega_r(\phi)\,\mathbf{u}_r(\boldsymbol{\mu}_n),
\label{eq:low-rank-ffd}
\end{equation}
where $\mathbf{u}_r\colon\mathbb{R}^3\to\mathbb{R}^{10}$ is the $r$-th spatial field, a cubic B-spline free-form deformation with control-point offsets in $\mathbb{R}^{10}$, and $\omega_r(\phi)$ is its scalar temporal coefficient.

\paragraph{Periodic B-spline in phase.}
To make the motion \emph{a periodic function of the phase}, we define each $\omega_r$ through a periodic cubic B-spline in $\phi$ with $M$ uniformly spaced control values $h_{r,m}$ and circular indexing, as in the cyclic temporal B-splines of 4DCT registration~\cite{metz2011nonrigid,vandemeulebroucke2011spatiotemporal}.
Specifically,
\begin{equation}
\omega_r(\phi)=2\sigma\Big(\sum_{m=0}^{M-1}h_{r,m}\,B\big(M\phi-m\big)\Big)-1,
\label{eq:periodic-bspline}
\end{equation}
where $B(x)=\sum_{k\in\mathbb{Z}}B_3(x-kM)$ wraps the uniform cubic B-spline $B_3$ with period $M$, and $\sigma$ is the sigmoid.
We optimize the control-point offsets of all $\mathbf{u}_r$ and the control values $h_{r,m}$ together with $\mathcal{G}_0$.
DIGS also defines its temporal coefficients with B-splines, but on \emph{acquisition time} and without periodicity.
X$^2$-Gaussian~\cite{yu2025x2gaussian} encourages periodicity only through a loss, which does not guarantee it.
Ours is periodic by construction.

\subsection{Implementation details}
\label{sec:projection-priors}
\label{sec:projection-optimization}

For training the Gaussian representation, we densify and prune the Gaussians as in 3DGS~\cite{kerbl2023gaussian} and X$^2$-Gaussian~\cite{yu2025x2gaussian}.
At each training iteration, we render $K$ random training projections at their respiratory phases, so that every update sees several gantry angles.

\paragraph{Initialization.}
Following prior work~\cite{zha2024r2gaussian, yu2025x2gaussian}, we initialize the Gaussians from a static volume $V_0$ that we reconstruct from the training projections with FDK in TIGRE~\cite{biguri2016tigre}.
$V_0$ approximates the time-averaged anatomy.
To focus the initial Gaussians on the body, we find the body by thresholding $V_0$~\cite{zha2024r2gaussian}, with a threshold that we choose by Otsu's method~\cite{otsu1979threshold} instead of a fixed value.
To place more Gaussians on edges~\cite{fu2025code}, we sample the initial centers inside the body from a mixture of two distributions, one proportional to the gradient magnitude $\|\nabla V_0\|_2$ with weight $\eta$, and one uniform with weight $1-\eta$.
We set the initial density of each Gaussian to $\kappa_{\rho}V_0(\boldsymbol{\mu}_n)$ with a scale factor $\kappa_{\rho}$, as in R$^2$-Gaussian~\cite{zha2024r2gaussian}.
Interestingly, we find that starting from more Gaussians than X$^2$-Gaussian++'s default significantly increases accuracy (\cref{sec:ablations}).
Increasing initial Gaussians near edges also helps edge clarity (\cref{sec:ablations}).

\paragraph{Weights of $\mathcal{L}_{\mathrm{wTV}}$.}
To weaken TV on edges (\cref{eq:wtv}), we find the edges by Otsu thresholding~\cite{otsu1979threshold} of $\|\nabla V_0\|_2$ inside the body (see the \suppmat).
We set the weight of a voxel to $\beta{=}0.3$ on these edges, to 0 outside the body, and to 1 elsewhere.
We compute $\nabla V_0$ and the differences in \cref{eq:wtv} with voxel-grid finite differences, without spacing normalization.

\paragraph{Hyperparameter selection.}
We sweep the hyperparameters on the three SPARE~\cite{shieh2018spareaapm,shieh2019spare} cases and choose the values with the highest body PSNR.
We then use these values unchanged for all our experiments.

\paragraph{Settings.}
We use the default settings of X$^2$-Gaussian++~\cite{yu2025x2gaussian,x2gaussianpp2026code}: the Adam optimizer~\cite{kingma2015adam}, $\lambda_{\mathrm{s}}=0.25$, $\lambda_{\mathrm{w}}=10$, $\lambda_{\mathrm{TV}}=0.05$, one $32^3$ TV patch per iteration, $\kappa_{\rho}=0.15$, an 8-voxel spatial control spacing, and its densification and pruning schedule.
We change the following: we start from 200K Gaussians instead of 50K, and use rank $R=8$.
As our method converges within 13K iterations, we run 13K iterations instead of 30K to save compute.
For parameters specific to our method,
we use $\eta=0.75$, $M=48$, and $K=4$, and we place each TV patch inside the body at the phase of one of the $K$ sampled projections.

\paragraph{Runtime and memory.}
On a single NVIDIA GeForce RTX 5090 GPU, the mean training time is 40\,min for our method and 37\,min for X$^2$-Gaussian.
Our method uses at most 30\,GiB of GPU memory during training.

\section{Experiments}
\label{sec:experiments}

\subsection{Experimental setup}
\label{sec:experimental-setup}

\paragraph{Datasets and preprocessing.} We use the same 13 cases as X$^2$-Gaussian~\cite{yu2025x2gaussian}: five DIR-Lab (DIR)~\cite{castillo2009framework}, five 4D-Lung~\cite{hugo2016lung,hugo2017longitudinal,clark2013tcia}, and three SPARE cases~\cite{shieh2018spareaapm,shieh2019spare}.\footnote{%
All cases are public and de-identified.
The SPARE dataset is licensed under the CC BY-NC 4.0 license (\url{https://creativecommons.org/licenses/by-nc/4.0/}).}
As noted earlier, we select our hyperparameters on the three SPARE cases, and test on the other 10.

For synthetic data, following X$^2$-Gaussian, we simulate 400-view, 60\,s scans with TIGRE~\cite{biguri2016tigre}, and use 300 views for training and 100 for held-out evaluation.
For realistic breathing, we deform the end-exhale CT of each case with a PCA motion model of its 10 phases, as in DIGS~\cite{huang2025digs}.
We vary the amplitude of each breathing cycle, add a slow drift, and add noise to the projections (see the \suppmat).
As the breathing rate varies within clinical scans~\cite{rijken2023individualized}, we draw the length of each cycle uniformly from 2.4 to 3.6\,s.
To keep one attenuation scale for all phases, as in a real scan, we divide all volumes of a patient by their common maximum.

We also reconstruct two real patient scans from the clinical sets of SPARE~\cite{shieh2018spareaapm,shieh2019spare}.
As these scans have no ground-truth volume, we compare them only qualitatively.

\begin{figure*}[t]
\centering
\includegraphics[width=\textwidth]{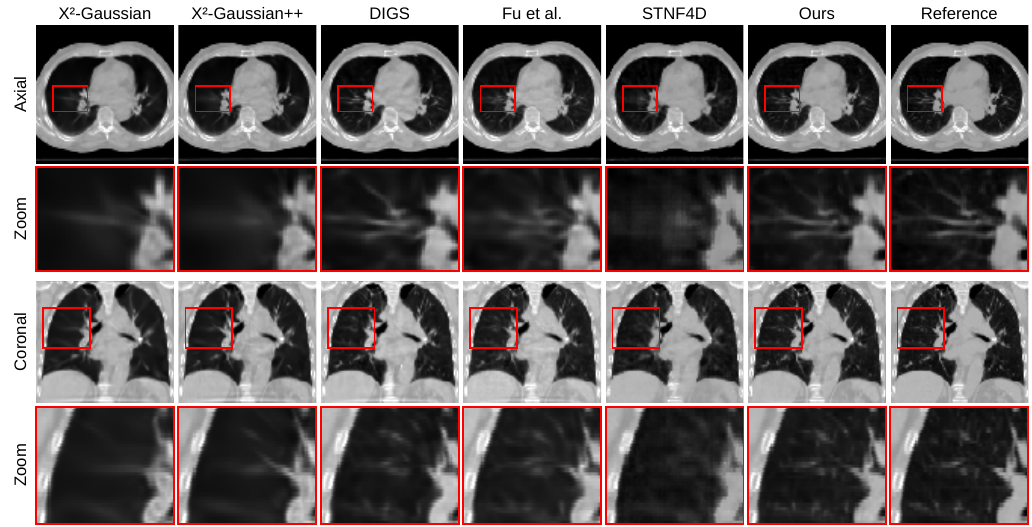}
\caption{\textbf{Qualitative comparison.}
The first and third rows show axial and coronal slices of DIR Case 3 with irregular breathing, and the second and fourth rows show the zoom-ins.
All methods use the same time point, slice locations, and display windows.
Our method provides the best reconstruction with fine pulmonary structures.
}
\label{fig:qualitative-comparison}
\end{figure*}

\paragraph{Baselines.}
We compare with two groups of dynamic methods and with conventional methods.
The first group defines its motion on acquisition time.
X$^2$-Gaussian++~\cite{yu2025x2gaussian,x2gaussianpp2026code}, the state of the art, and its predecessor X$^2$-Gaussian~\cite{yu2025x2gaussian} learn the breathing period.
DIGS~\cite{huang2025digs} uses the B-spline motion that we build on.
The second group ties each projection to a phase bin.
Fu \etal~\cite{fu2025dynamiccbct} deform one anatomy with a motion for each bin, and STNF4D~\cite{zhou2025stnf4d} is a neural field that takes the bin index as input.
As these two methods need the bin of each projection, we give them \emph{the ground-truth phase bins}.
To show how accurate reconstruction is without a learned motion, we also compare with conventional static methods.
We reconstruct 10 bins of our surrogate phase and one static volume, each with FDK~\cite{feldkamp1984practical} and with conjugate gradient least squares (CGLS)~\cite{hestenes1952methods}, and run PICCS~\cite{chen2008piccs} with our $V_0$ as its prior image.
To measure what our surrogate phase costs, we also run our method with the ground-truth phase.
We run every compared method on the same simulated projections, and the dynamic baselines with the settings released by their authors (see the \suppmat).\footnote{We do not compare with other recent dynamic methods~\cite{shao2024pmfstinr,xie2025pmfstgr,zhou2026adiner,zhang2026dnaf,zhong2026tgfield} or motion-compensated methods~\cite{wang2013smeir,huang2024surrogate}, because their reconstruction code is not public at the time of writing.}

\paragraph{Evaluation metrics.}
\label{sec:evaluation-protocol}
As the breathing changes over the scan, we evaluate the reconstructions at 10 time points in each of four complete breathing cycles of every 60\,s scan: the first, middle, and last cycles, and the longest one.
To evaluate the volume, we report PSNR, SSIM, and the high-frequency error norm (HFEN)~\cite{ravishankar2011mr} over the whole volume and inside the body, which we extract from each reference volume (see the \suppmat).
HFEN measures the error of fine structures such as vessels and airway walls.
To evaluate how well each method predicts the 100 held-out projections, we report their PSNR and SSIM as well.
To evaluate motion, we report the motion error, the mean distance between the predicted and simulated displacements of 1,024 random points inside the body, measured from the first end-exhale time point.
On DIR, we also report the target registration error (TRE) on the 75 landmarks~\cite{castillo2009framework,dirlab4dct} that experts annotated in the six phases from end-inhale (T00) to end-exhale (T50).
See the \suppmat{} for details.

\subsection{Comparison with the state of the art}
\label{sec:main-results}

\begin{table*}[t]
\caption{\textbf{Comparison with the state of the art.}
We average the values over the 10 DIR and 4D-Lung cases and all evaluated time points.
We mark methods that use ground-truth phase bins with $^{*}$.
We also provide the baseline motion error assuming static reconstruction.
We mark the best values of each column in bold, excluding our method with the ground-truth phase, which we provide as reference.
Our method provides the best performance on \emph{every} metric.
}
\label{tab:main-results}
\vspace{-1em}
\begin{center}
\resizebox{\textwidth}{!}{%
\setlength{\tabcolsep}{12pt}%
\begin{tabular}{@{}lccccccccc@{}}
\toprule
& \multicolumn{3}{c}{Body volume}
& \multicolumn{3}{c}{Whole volume}
& \multicolumn{2}{c}{Held-out projection}
& Motion \\
\cmidrule(lr){2-4}
\cmidrule(lr){5-7}
\cmidrule(lr){8-9}
Method
& PSNR (dB)$\uparrow$ & SSIM$\uparrow$ & HFEN$\downarrow$
& PSNR (dB)$\uparrow$ & SSIM$\uparrow$ & HFEN$\downarrow$
& PSNR (dB)$\uparrow$ & SSIM$\uparrow$ & error (mm)$\downarrow$ \\
\midrule

Phase-binned FDK~\cite{feldkamp1984practical}\tabvenue{JOSA A'84}
& 24.63 & 0.5350 & 1.4193
& 23.62 & 0.3506 & 1.9280
& 24.16 & 0.7934 & - \\

Phase-binned CGLS~\cite{hestenes1952methods}\tabvenue{J. Res. NBS'52}
& 29.99 & 0.8095 & 0.7541
& 31.08 & 0.6559 & 0.8808
& 40.19 & 0.9655 & - \\

Static FDK~\cite{feldkamp1984practical}\tabvenue{JOSA A'84}
& 31.59 & 0.9137 & 0.3990
& 33.13 & 0.7352 & 0.4488
& 28.94 & 0.8849 & - \\

Static CGLS~\cite{hestenes1952methods}\tabvenue{J. Res. NBS'52}
& 36.73 & 0.9324 & 0.3938
& 38.63 & 0.8984 & 0.4167
& 42.23 & 0.9865 & - \\

PICCS~\cite{chen2008piccs}\tabvenue{Med. Phys.'08}
& 39.40 & 0.9516 & 0.3172
& 41.38 & 0.9616 & 0.3235
& 48.00 & 0.9968 & - \\

\midrule

\textcolor{gray}{Static}
& \textcolor{gray}{-} & \textcolor{gray}{-} & \textcolor{gray}{-}
& \textcolor{gray}{-} & \textcolor{gray}{-} & \textcolor{gray}{-}
& \textcolor{gray}{-} & \textcolor{gray}{-} & \textcolor{gray}{1.16} \\

X$^2$-Gaussian~\cite{yu2025x2gaussian}\tabvenue{ICCV'25}
& 36.94 & 0.9421 & 0.3869
& 39.84 & 0.9664 & 0.3535
& 48.63 & 0.9955 & 1.11 \\

X$^2$-Gaussian++~\cite{x2gaussianpp2026code}\tabvenue{GitHub'26}
& 38.24 & 0.9570 & 0.3255
& 41.32 & 0.9743 & 0.2894
& 48.90 & 0.9957 & 1.08 \\

DIGS~\cite{huang2025digs}\tabvenue{MICCAI'25}
& 38.01 & 0.9638 & 0.2895
& 40.76 & 0.9735 & 0.2789
& 51.35 & 0.9965 & 1.54 \\

Fu \etal$^{*}$~\cite{fu2025dynamiccbct}\tabvenue{Med. Phys.'25}
& 37.70 & 0.9664 & 0.2694
& 41.03 & 0.9798 & 0.2340
& 51.31 & 0.9979 & 0.97 \\

STNF4D$^{*}$~\cite{zhou2025stnf4d}\tabvenue{AAAI'25}
& 38.57 & 0.9392 & 0.3308
& 39.94 & 0.9568 & 0.3671
& 48.11 & 0.9943 & - \\

Ours
& \textbf{42.18} & \textbf{0.9755} & \textbf{0.1868}
& \textbf{44.44} & \textbf{0.9834} & \textbf{0.1887}
& \textbf{52.04} & \textbf{0.9980} & \textbf{0.57} \\

\addlinespace[3pt]

\textcolor{gray}{Ours (GT phase)}
& \textcolor{gray}{42.25} & \textcolor{gray}{0.9756} & \textcolor{gray}{0.1857}
& \textcolor{gray}{44.30} & \textcolor{gray}{0.9831} & \textcolor{gray}{0.1938}
& \textcolor{gray}{52.08} & \textcolor{gray}{0.9980} & \textcolor{gray}{0.56} \\

\bottomrule
\end{tabular}}

\parbox{\textwidth}{\footnotesize
-: not applicable, as the conventional baselines and STNF4D estimate no motion, and Static gives no volume.}
\end{center}
\end{table*}

\begin{table}[t]
\caption{\textbf{Landmark and volume accuracy on real anatomy.}
We also report the tracking performance (TRE) of expert-annotated landmarks of the five DIR cases, simulated from their real phase volumes.
We also report the body PSNR averaged over the 10 phases of the first cycle.
We mark methods that use ground-truth phase bins with $^{*}$.
We also report the error which a static model would make as reference.
Our method tracks the landmarks the best, which also leads to better body PSNR.
}
\label{tab:tre}
\vspace{-1em}
\begin{center}
\resizebox{\linewidth}{!}{%
\setlength{\tabcolsep}{3pt}%
\begin{tabular}{@{}lcccccc@{}}
\toprule
& \shortstack{Static\\Recon.}
& \shortstack{X$^2$-Gaussian\\\cite{yu2025x2gaussian}\tabvenue{ICCV'25}}
& \shortstack{X$^2$-Gaussian++\\\cite{x2gaussianpp2026code}\tabvenue{GitHub'26}}
& \shortstack{DIGS\\\cite{huang2025digs}\tabvenue{MICCAI'25}}
& \shortstack{Fu \etal$^{*}$\\\cite{fu2025dynamiccbct}\tabvenue{Med. Phys.'25}}
& Ours \\
\midrule
TRE (mm)$\downarrow$ & 4.58 & 4.34 & 4.07 & 3.33 & 3.84 & 1.82 \\
Body PSNR (dB)$\uparrow$ & - & 39.44 & 40.62 & 39.49 & 41.09 & 44.16 \\
\bottomrule
\end{tabular}}
\end{center}
\end{table}

\paragraph{Qualitative comparison.}
Our method provides the best reconstructions with sharp fine structures as shown in \cref{fig:qualitative-comparison}.
Notice how the small pulmonary vessels are either blurred or missing in other baselines.
As these structures cover few voxels, quantitative metrics such as PSNR over the whole body hide our gain.
This gain, however, is visible via body HFEN, where our method is $\approx$30\% below that of the next-best method in \cref{tab:main-results}.

\paragraph{Quantitative comparison.}
The qualitative gains we show in \cref{fig:qualitative-comparison} are confirmed quantitatively in \cref{tab:main-results}.
We achieve the best performance in \emph{every metric}.
Our body PSNR exceeds that of every dynamic baseline by more than 3\,dB, even those of Fu \etal and STNF4D, which use the ground-truth phase bins.
Our motion error is about half that of the static baseline, and 0.40\,mm below that of Fu \etal, the next best.
We analyze where these gains come from in \cref{sec:ablations}.

\paragraph{Landmarks on real anatomy.}
As we report in \cref{tab:tre}, our method also provides the most accurate tracks for the expert-annotated landmarks of DIR.
Because the landmarks are annotated in the real phase volumes, we compute TRE on DIR scans simulated from these volumes (note that these simulations have varying cycle lengths but no amplitude variation, drift, or noise).
As shown, accurate tracks also lead to accurate body PSNR, exceeding those of the other methods by more than 3\,dB.

\begin{figure*}[t]
\centering
\includegraphics[width=\linewidth]{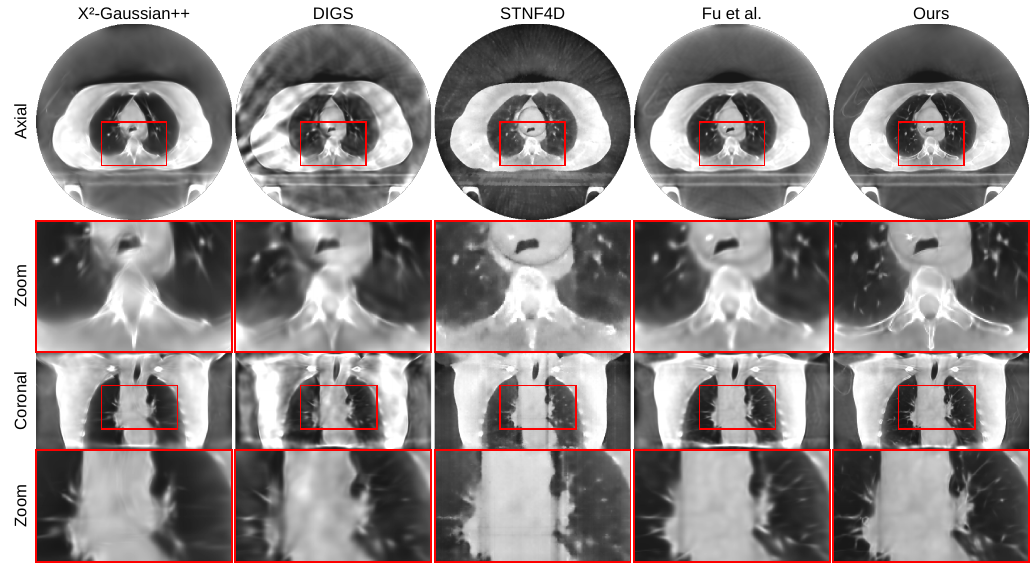}
\caption{\textbf{Qualitative comparison on a real patient scan.}
We show the reconstruction outcome for SPARE (Varian patient 1, phase bin 06).
The first and third rows show axial and coronal slices, and the others are the zoom-ins.
All columns use the same slice locations and display windows.
Fu \etal and STNF4D are provided with the SPARE phase bins.
Our method clearly reconstructs the pulmonary vessels and the bone edges, while other methods show lower quality.
}
\label{fig:clinical}
\end{figure*}

\paragraph{Real patient scans.}
To remove any potential simulation biases, we further use real patient scans in \cref{fig:teaser,fig:clinical}.
As these scans do not have ground-truth volumes, we examine them qualitatively.
Our method shows clear reconstructions and demonstrates a similar gap in reconstruction quality as in simulated data.

\begin{table}[t]
\caption{\textbf{Ablation study.}
We compare with various modifications of our method on the synthetic data.
Accurate grounding is critical to the best performance.
Using one value per phase bin to represent motion provides slight degradation, but results in abrupt motion as bins are discontinuous.
Implementation choices also provide minor gains.
Edge PSNR is the PSNR of voxels within two voxels of edges.
}
\label{tab:ablations}
\vspace{-1em}
\begin{center}
\resizebox{\linewidth}{!}{%
\setlength{\tabcolsep}{4pt}%
\begin{tabular}{@{}lcccc@{}}
\toprule
& \multicolumn{3}{c}{Body volume} & Motion \\
\cmidrule(lr){2-4}
Variant & PSNR (dB)$\uparrow$ & HFEN$\downarrow$ & Edge PSNR (dB)$\uparrow$ & error (mm)$\downarrow$ \\
\midrule
\textbf{Our full method }& 42.18 & 0.1868 & 40.01 & 0.57 \\
\midrule
\multicolumn{5}{l}{\emph{Grounding}} \\[2pt]
W/o grounding (on acquisition time) & 37.63 & 0.2988 & 34.87 & 0.80 \\
\quad + periodic consistency loss & 37.76 & 0.3367 & 34.30 & 1.04 \\
\addlinespace[2pt]
\midrule
\multicolumn{5}{l}{\emph{Accuracy of the phase}} \\[2pt]
Learned period w/o $\delta(t)$ & 39.48 & 0.2742 & 36.45 & 0.81 \\
W/o $\delta(t)$ & 39.38 & 0.2777 & 36.39 & 0.81 \\
\addlinespace[2pt]
\midrule
\multicolumn{5}{l}{\emph{Sharing one motion}} \\[2pt]
One value per phase bin & 42.16 & 0.1878 & 39.94 & 0.57 \\
Separate motion per phase bin & 39.44 & 0.2456 & 37.05 & 0.88 \\
\addlinespace[2pt]
\midrule
\multicolumn{5}{l}{\emph{Implementation choices}} \\[2pt]
50K initial Gaussians & 41.35 & 0.2183 & 39.13 & 0.62 \\
Uniform initial sampling ($\eta{=}0$) & 42.08 & 0.1874 & 39.90 & 0.57 \\
Uniform TV ($\beta{=}1$) & 42.06 & 0.1904 & 39.85 & 0.57 \\
One view per update ($K{=}1$) & 41.50 & 0.2002 & 39.69 & 0.56 \\
\bottomrule
\end{tabular}}
\end{center}
\end{table}

\subsection{Ablation studies}
\label{sec:ablations}

To show how much each component of our method contributes, we perform an ablation study on the synthetic data.
We report our results in \cref{tab:ablations}.

\paragraph{Grounding.}
Grounding the motion in an accurate phase is critical to the method.
When we define our temporal B-splines on acquisition time instead, as DIGS does, body PSNR drops by 4.55\,dB.
Projections from different breathing cycles then no longer constrain the same motion.
Enforcing periodic consistency~\cite{yu2025x2gaussian,x2gaussianpp2026code} does not help and rather harms, raising the motion error from 0.80 to 1.04\,mm.
As the amplitude and drift differ from one breathing cycle to the next, the motion likely blurs out to satisfy the consistency loss.
Asking the motion to repeat is thus no substitute for grounding it in the phase.

\paragraph{Accuracy of the phase.}
Grounding loses most of its gain with an inaccurate phase.
When we learn one breathing period instead of reading the phase from the surrogate, the phase drifts away from the breathing.
The volume accuracy then drops, but stays above that of the variant without grounding.
When we remove only $\delta(t)$, the correction in \cref{eq:phase-coordinate} that lets the phase advance faster or slower, the phase cannot follow cycles of different lengths.

\paragraph{Sharing one motion.}
While using one value per phase bin causes a minor degradation in performance, the reconstructions show abrupt motion as bins are discrete.
Our periodic B-splines allow motion to be smooth in phase by construction, yielding the best performance.

\paragraph{Implementation choices.}
We further outlined other minor components of our method.
Using fewer initial Gaussians lowers body PSNR by 0.83\,dB, and updating one view at a time also reduces it by 0.68\,dB.
Uniform initial sampling and uniform TV both slightly lower the volume accuracy.

\section{Conclusion}
\label{sec:conclusion}

We proposed BreathSplat, which grounds one smooth motion of the anatomy in the respiratory phase.
To do so, we move one set of Gaussians with B-splines that are periodic in the phase, which we read from conventional surrogates.
The projections of every breathing cycle thus inform one shared anatomy and motion.
With simulated scans, we showed that our method outperforms the state of the art.
With two real patient scans, we further showed that our method reconstructs fine structures more clearly than the dynamic baselines.
We further provided an ablation study that demonstrates the importance of accurate phase-based grounding.

\paragraph{Limitations and future work.}
To let every projection inform the shared anatomy at every phase, we assume that the phase alone determines the motion.
As we model the motion as a smooth function of the phase, we also cannot follow abrupt changes, such as the change of speed at end-inhale that Vandemeulebroucke \etal~\cite{vandemeulebroucke2011spatiotemporal} found in 4DCT.
While our real-patient scan results show that this is not a major problem in typical 4D-CBCT, it remains an interesting avenue of study.
For example, we could treat the motion that the phase and a smooth motion do not explain, including abrupt changes, as noise instead of modeling it, and make the reconstruction robust to it, as robust losses~\cite{sabour2023robustnerf,sabour2025spotlesssplats} do for transient distractors in radiance fields and Gaussian splatting.

\section*{Acknowledgments}
This work was supported in part by the Natural Sciences and Engineering Research Council of Canada (NSERC) Discovery Grant, Digital Research Alliance of Canada, and Advanced Research Computing at the University of British Columbia.
This research was also supported by the MSIT(Ministry of Science and ICT), Korea, under the Global Research Support Program in the Digital Field program(IITP-2026-RS-2024-00418641) and the ITRC(Information Technology Research Center) support program(IITP-2026-RS-2024-00437102) supervised by the IITP(Institute for Information \& Communications Technology Planning \& Evaluation).

{
    \small
    \bibliographystyle{ieeenat_fullname}
    \bibliography{main}
}
\clearpage
\appendix
\maketitlesupplementary

\section{Additional qualitative comparisons}
\label{sec:suppl-qualitative}

\begin{figure*}[t]
\centering

\textbf{(a) 4D-Lung Case 4}\par
\includegraphics[width=0.92\textwidth]{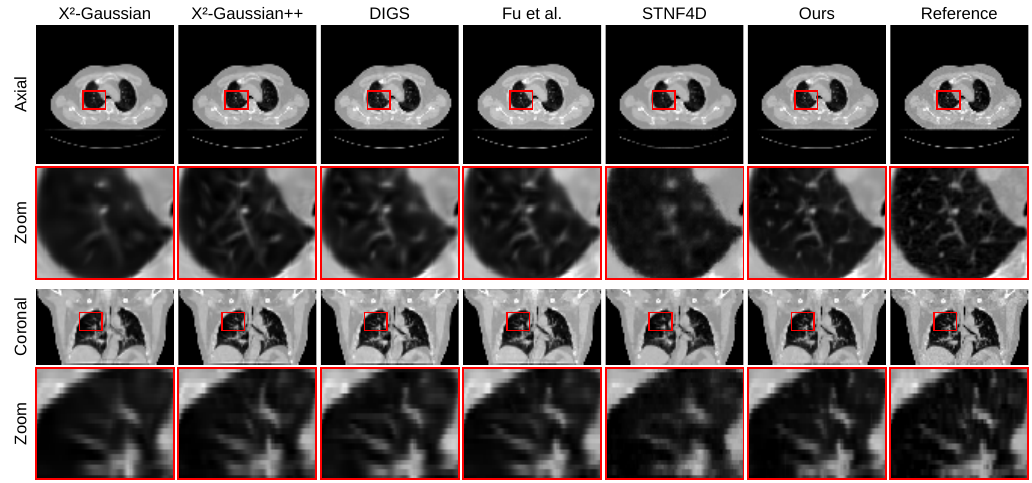}

\vspace{0.6em}

\textbf{(b) SPARE Case 3}\par
\includegraphics[width=0.92\textwidth]{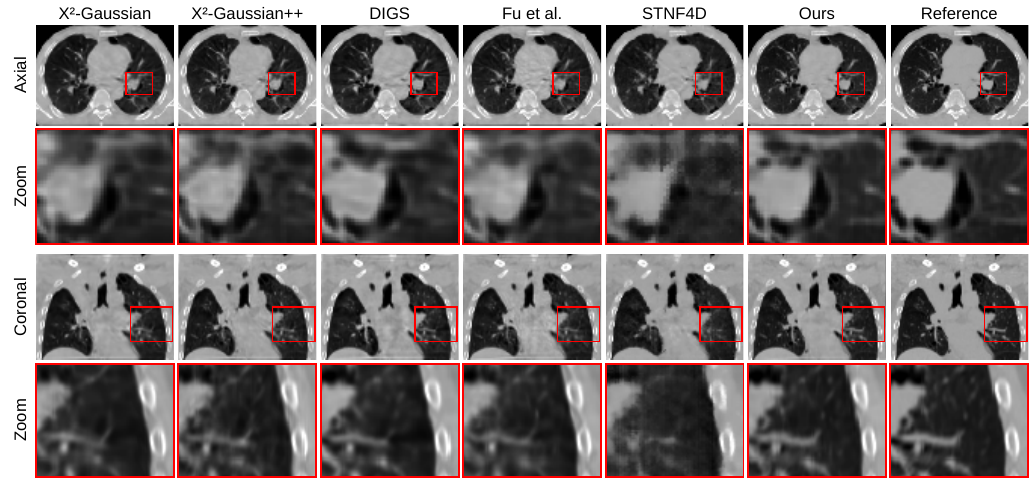}

\caption{\textbf{Qualitative comparison on 4D-Lung and SPARE.}
The first and third rows of each panel show axial and coronal slices of (a) 4D-Lung Case 4 and (b) SPARE Case 3, both simulated with irregular breathing, and the second and fourth rows show the zoom-ins.
All methods within each case use the same time point, slice locations, and display windows.
Our method provides the best reconstruction with fine pulmonary structures.}

\label{fig:suppl-qualitative-comparison}
\end{figure*}

Our method also provides the best reconstructions with sharp fine structures on 4D-Lung and SPARE, as shown in \cref{fig:suppl-qualitative-comparison}.
Notice how the small pulmonary vessels are either blurred or missing in the dynamic baselines.

\section{Accuracy of the phase}
\label{sec:suppl-fixed-c}
\label{sec:suppl-variable-ablation}

Thanks to the correction $\delta(t)$ in \cref{eq:phase-coordinate}, our surrogate phase stays close to the ground-truth phase.
With irregular breathing, the surrogate phase differs from the ground-truth phase by a root-mean-square error of 0.015 cycles, and by 0.104 cycles without $\delta(t)$.
We average this error over the 10 DIR and 4D-Lung cases and the three SPARE cases.

\section{Evaluation metrics}
\label{sec:suppl-metrics}

Following the 10-phase evaluation of X$^2$-Gaussian~\cite{yu2025x2gaussian}, we evaluate the continuous-time reconstructions on the scans of \cref{tab:tre} at normalized phase-center times $\tau_k$, $k=0,\ldots,9$, within the first breathing cycle.
Each $\tau_k$ is the normalized time at which the ground-truth phase reaches $(k+0.5)/10$.
On the other simulated scans, we evaluate the reconstructions at the time points of the same 10 phases in each of the four cycles of \cref{sec:evaluation-protocol}.
We obtain the reference volume of these scans at each time point by deforming the end-exhale CT with the PCA motion model at that time.

Volumetric PSNR uses the maximum of the reference volume as its peak and computes mean squared error within the evaluated region.
We average SSIM over slices in the three orthogonal orientations.
Body SSIM includes only windows contained within the body mask.
For HFEN, we fix the Laplacian-of-Gaussian width to 1.8\,mm across methods and datasets to assess fine-structure errors at a common physical scale.
For edge PSNR, we find the edges of each reference volume by Otsu thresholding of its gradient magnitude, as for $V_0$ (\cref{sec:projection-priors}).
Held-out projection PSNR and SSIM compare the renderings with the noise-free projections.

To compute TRE, we measure each predicted displacement relative to the deformation at T00, as we do not anchor the canonical Gaussians to T00 or any other respiratory phase.
Let $\mathbf{d}(\mathbf{x},\phi)$ denote the positional deformation of a point $\mathbf{x}$ at phase $\phi$, and let $\mathbf{x}_{\ell,0}$ and $\mathbf{x}_{\ell,k}$ denote the reference positions of landmark $\ell$ at T00 and at the later phase $k$.
We define the predicted displacement as $\widehat{\mathbf{v}}_{\ell,k}=\mathbf{d}(\mathbf{x}_{\ell,0},\phi(\tau_k))-\mathbf{d}(\mathbf{x}_{\ell,0},\phi(\tau_0))$.
For each baseline that models deformation, we query its deformation at the same time points through the time or phase input of its released code.
We likewise subtract its deformation at $\tau_0$.
The per-case TRE is
\begin{equation}
\operatorname{TRE}=
\frac{1}{5\cdot75}
\sum_{k=1}^{5}\sum_{\ell=1}^{75}
\left\|
\widehat{\mathbf{v}}_{\ell,k}
-(\mathbf{x}_{\ell,k}-\mathbf{x}_{\ell,0})
\right\|_2.
\label{eq:tre}
\end{equation}
The indices $k=1,\ldots,5$ denote T10 to T50, and all coordinates and displacements are in mm.

For the motion error, we sample the 1,024 points uniformly at random inside the body of each case, query the deformations as for TRE, and measure the displacements from the first end-exhale time point.
These points and the simulated motion differ from those of TRE, so the two errors are not comparable.

\begin{figure*}[t]
\centering
\includegraphics[width=\textwidth]{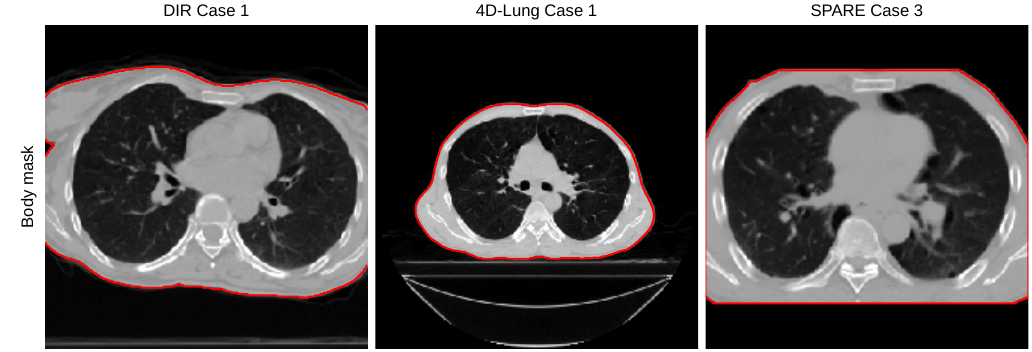}
\caption{\textbf{Evaluation-only body masks.}
We show the masks for DIR Case 1, 4D-Lung Case 1, and SPARE Case 3.
These masks are separate from the body region that we take from $V_0$.}
\label{fig:suppl-body-mask}
\end{figure*}

\paragraph{Evaluation-only body masks.}
\label{sec:suppl-body-mask}
To obtain the body mask of each reference volume, we first remove structures outside the patient by adapting the surface-detection strategy of Zhu \etal~\cite{zhu2012table}.
For DIR and 4D-Lung, we then extract the body with a $-500$\,Hounsfield unit (HU) threshold, as in CT body-contour extraction~\cite{maspero2018dose}.
For SPARE, we extract the body by thresholding the supplied attenuation values at $0.003\,\mathrm{mm}^{-1}$.
In all datasets, we then fill the inside of the body contour, so that the masks include the lungs.
\Cref{fig:suppl-body-mask} shows examples of the evaluation-only body masks of \cref{sec:evaluation-protocol}.

\section{Baselines}
\label{sec:suppl-setup}

For conventional comparisons, we reconstruct 10 phase bins and one static volume using FDK and 20-iteration CGLS, with bins assigned from the surrogate phase of the 300 training projections.
Bin $k$ holds the training projections with surrogate phase in $[k/10,(k+1)/10)$.
As the surrogate and ground-truth phases are both zero at scan onset, we compare bin $k$ with the $k$-th evaluation time point of each cycle.
We render each held-out projection from the bin that contains its surrogate phase.
The static FDK reconstruction is the volume $V_0$ that guides our method.
PICCS starts from $V_0$, runs at most 100 iterations per bin, and uses a prior weight of 0.5, which we select on the three SPARE cases.

Our method renders about as many projections as X$^2$-Gaussian and X$^2$-Gaussian++.
Our method uses 13K optimization iterations with four full projections per iteration, totaling 52K projection renderings.
X$^2$-Gaussian and X$^2$-Gaussian++ use 30K iterations, with 30K observed-view renderings and 23K additional renderings for periodic consistency, totaling 53K projection renderings.
DIGS uses 50K iterations, and Fu \etal use 30K iterations, both with one projection per iteration.
STNF4D uses 40K iterations with 1,024 sampled rays per iteration.
In the simulations, Fu \etal and STNF4D receive the ground-truth phase divided into 10 bins.

\paragraph{Periodic consistency loss.} For the periodic consistency loss that we add to the variant without grounding in \cref{tab:ablations}, we use the loss of the public X$^2$-Gaussian++ code~\cite{x2gaussianpp2026code} with its default weights, and apply it after the first 7K iterations, as that code does.
It renders each training view at the time one cycle of the surrogate phase away and compares the rendering with the measured projection.
With four views per iteration, this loss adds 24K renderings to the 52K observed-view renderings, totaling 76K projection renderings.

\section{Real patient scans}
\label{sec:suppl-clinical}

We use the first scans of patients 1 and 3 in the clinical Varian dataset of SPARE~\cite{shieh2018spareaapm,shieh2019spare}, both 8-min half-fan CBCT scans.
All methods in \cref{fig:teaser,fig:clinical} reconstruct each scan from the 680 equispaced projections that SPARE~\cite{shieh2019spare} extracts from its fully sampled projection set as the equivalent of a 1-min scan.
All use the same reconstruction settings as in the simulated experiments.
Fu \etal and STNF4D receive the phase bin that SPARE provides for each projection.
We show the reconstructions at the same acquisition time in phase bin 06 of SPARE for patient 1 (\cref{fig:clinical}) and in phase bin 01 for patient 3 (\cref{fig:teaser}), and those of Fu \etal and STNF4D at the index of each bin.

\section{Simulation of irregular breathing}
\label{sec:suppl-perturbed}

We follow the simulation of DIGS~\cite{huang2025digs}.
Unlike DIGS, we do not use sliding registration.
We set the slow drift, whose size DIGS does not specify, to 5\% of the range of the first PCA coefficient over the scan.
To add the projection noise in physical units, we undo the division by the common maximum (\cref{sec:experimental-setup}) and multiply the attenuation of DIR and 4D-Lung relative to water by the attenuation coefficient of water at 75\,keV, $0.01879\,\mathrm{mm}^{-1}$~\cite{berger2010xcom}.

\section{Additional implementation details}

\label{sec:suppl-phase-fitting}To fit the phase of \cref{eq:phase-coordinate}, we write the repeating waveform of \cref{sec:respiratory-timing} as a Fourier series.
We initialize $c$ with the Lomb--Scargle periodogram~\cite{lomb1976least,scargle1982studies}.
We then fit $c$ and $\delta(t)$, and obtain the sine and cosine coefficients for the current $c$ and $\delta(t)$ by linear least squares~\cite{golub1973differentiation}.

\begin{figure*}[t]
\centering

\includegraphics[width=\textwidth]{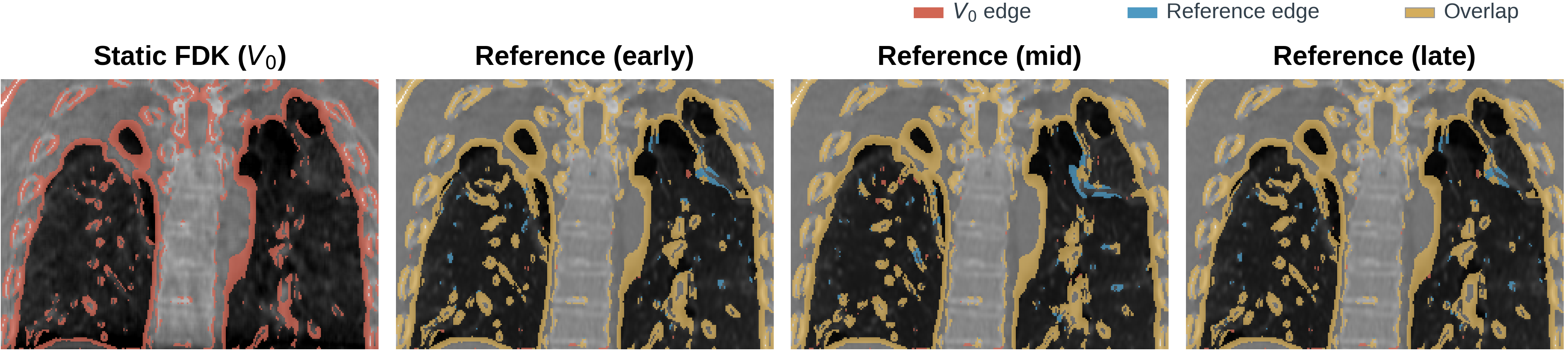}

\caption{\textbf{Edges of $V_0$ and of the reference volumes.}
Comparison of the edge map from the static FDK reconstruction $V_0$ with those from the reference volumes (\cref{sec:experimental-setup}) at three time points in one breathing cycle of simulated SPARE Case 3.
Most edges of $V_0$ lie on edges of the reference volumes at all three time points.}
\label{fig:edge-guidance}
\end{figure*}

\paragraph{Weights of $\mathcal{L}_{\mathrm{wTV}}$.} \Cref{fig:edge-guidance} compares the edges that we find in $V_0$ with those of the reference volumes.

\section{Hyperparameter sensitivity}
\label{sec:suppl-parameter-selection}

\Cref{tab:suppl-parameter-sweep} shows the sensitivity of our method to its hyperparameters on DIR and 4D-Lung.

\begin{table*}[t]
\caption{\textbf{Hyperparameter sensitivity.}
We report the changes in body metrics on the 10 DIR and 4D-Lung cases relative to the selected values.
We select the values on the three SPARE cases.
All values of $K$ use the same number of projection renderings.
On DIR and 4D-Lung, the selected values are within 0.05\,dB in body PSNR of the best value in each sweep.}
\label{tab:suppl-parameter-sweep}
\vspace{-1em}
\begin{center}
\resizebox{\textwidth}{!}{%
\setlength{\tabcolsep}{52pt}%
\begin{tabular}{@{}llrrr@{}}
\toprule
Parameter & Value & $\Delta$PSNR (dB)$\uparrow$ & $\Delta$SSIM$\uparrow$ & $\Delta$HFEN$\downarrow$ \\
\midrule
Rank $R$ & 2 & $-$0.03 & $-$0.0001 & +0.0007 \\
& 4 & +0.01 & $-$0.0001 & $-$0.0009 \\
& 8 (selected) & 0.00 & 0.0000 & 0.0000 \\
& 16 & $-$0.36 & $-$0.0008 & +0.0069 \\
\midrule
Temporal control values $M$ & 12 & $-$0.05 & 0.0000 & +0.0007 \\
& 24 & $-$0.02 & $-$0.0003 & +0.0001 \\
& 48 (selected) & 0.00 & 0.0000 & 0.0000 \\
& 96 & $-$0.09 & $-$0.0004 & +0.0020 \\
\midrule
Initial Gaussians & 50K & $-$0.83 & $-$0.0069 & +0.0315 \\
& 100K & $-$0.16 & $-$0.0014 & +0.0071 \\
& 200K (selected) & 0.00 & 0.0000 & 0.0000 \\
& 300K & $-$0.07 & +0.0009 & $-$0.0001 \\
\midrule
Sampling mixture $\eta$
& 0 (uniform)
& $-$0.10 & $-$0.0007 & +0.0006 \\
& 0.75 (selected)
& 0.00 & 0.0000 & 0.0000 \\
& 1.0
& +0.00 & 0.0000 & +0.0007 \\
\midrule
Views per update $K$ & 1 & $-$0.68 & $-$0.0021 & +0.0134 \\
& 2 & $-$0.11 & $-$0.0002 & +0.0005 \\
& 4 (selected) & 0.00 & 0.0000 & 0.0000 \\
& 8 & $-$0.24 & $-$0.0007 & +0.0071 \\
\midrule
Edge weight $\beta$ & 0 & +0.04 & +0.0001 & $-$0.0009 \\
& 0.1 & +0.02 & +0.0001 & $-$0.0004 \\
& 0.3 (selected) & 0.00 & 0.0000 & 0.0000 \\
& 1.0 (uniform TV) & $-$0.12 & $-$0.0004 & +0.0036 \\
\bottomrule
\end{tabular}}
\end{center}
\end{table*}

\end{document}